# DYNAMIC OBJECT COMPREHENSION:
# A FRAMEWORK FOR EVALUATING ARTIFICIAL VISUAL PERCEPTION

*Scott Y.L. Chin*[*τ,] *Bradley R. Quinton*[*τ]

[*]Singulos Research Inc.
[τ]Dept. of Electrical and Computer Engineering, University of British Columbia
Email: scottc@singulosresearch.com; bradq@singulosresearch.com

**ABSTRACT**

Augmented and Mixed Reality are emerging as likely successors to the mobile internet. However, many technical challenges remain. One of the key requirements of these systems is the ability to create a continuity between physical and virtual worlds, with the user's visual perception as the primary interface medium. Building this continuity requires the system to develop a visual understanding of the physical world. While there has been significant recent progress in computer vision and AI techniques such as image classification and object detection, success in these areas has not yet led to the *visual perception* required for these critical MR and AR applications. A significant issue is that current evaluation criteria are insufficient for these applications. To motivate and evaluate progress in this emerging area, there is a need for new metrics. In this paper we outline limitations of current evaluation criteria and propose new criteria.

## 1. INTRODUCTION

Mixed-Reality (MR) is an environment where physical and digital objects can co-exist and interact [1]. A possible criterion for an MR system is the ability for effortless integration of data between the physical and digital objects in either direction [2]. The MR applications that we currently see deployed in society typically fall into the category of Augmented Reality (AR) [3]. AR can be described as a point on the Mixed-Reality Spectrum [4].

Augmented Reality systems are commonly implemented on mobile handheld devices or head-mounted displays (HMD) [5]. The user is typically provided an egocentric view [6] of the physical world through a display on the device, and digital objects are overlayed onto this view in real-time as the user moves through the physical world. To align the digital objects with the physical world in this view, AR systems are configured with sensors to gather data regarding the physical environment. The most common sensors are video cameras and inertial sensors (accelerometers, gyroscopes, and magnetometers). Techniques such as Visual-Inertial Odometry [7] can then be used to determine the position and orientation of the observing view in the 3D physical world.

The MR/AR systems that we see today generally do not yet achieve the goal of effortless integration of data between the physical and digital worlds because they are not yet capable of acquiring a detailed understanding of their physical environment. In specific scenarios, they can align digital objects within the physical world, but only once physical objects or locations are first identified by the user. The resulting interactions between the worlds are typically very limited and quite static. In other words, these systems do not *understand* the world around them, instead they ask the MR user to explain it to them. In a move towards addressing this limitation, recent devices are beginning to include depth sensors (e.g., time-of-flight, lidar) that can help with and spatial mapping [8], and frameworks to help with structural scene understanding [9][10] (*e.g.*, the location of planes and surfaces). These advancements provide a coarse geometrical understanding of the physical environment.

To achieve effortless integration of data between the physical and digital worlds, MR systems require a real-time Semantic Comprehension of the physical objects in the physical environment. Semantic Comprehension includes, but is not limited to, understanding what an object is, where the object is in the physical environment, and an understanding of the properties of the object. Modern computer vision and machine learning techniques would seem to be able to serve as the backbone to Semantic Comprehension engines. These techniques include image classification [11-14], object detection [15-17], object tracking [18][19] and object segmentation [20-22].

However, there are notable omissions in the definition of the goal of these techniques, when applied to Mixed Reality. One is the consideration that the world and its objects are three-dimensional. The aforementioned techniques generally operate on 2D images – projections of the 3D world onto 2D sensors. Research has also focused on working with 3D data directly such point-clouds [23] and voxels [24]. But 3D data is not as ubiquitous nor as dense as visual data. A new approach which fundamentally acknowledges that we desire a Semantic Comprehension of 3D objects that exist in a 3D world *interpreted* through 2D data is desirable.

Furthermore, current evaluation metrics are insufficient to measure overall "visual perception" from a MR user's point of view. AI researchers report systems that surpass human-level accuracy [25] [26]. And yet we see few, if any, examples of robust deployments of MR systems that achieve real-time Semantic Comprehension in real-life scenarios. Clearly these metrics alone are not sufficient for evaluating or driving AI-based computer vision for these applications.

The key short comings are: 1) a failure to evaluate latency compared to human ability/expectation (*i.e.* how quickly do we need to get to an answer), 2) failure to evaluate 3D localization (*i.e.* understanding the location of an object in the world co-ordinate system rather than in the image), 3) failure to account for object types that don't need to be supported by the system, and 4) failure to consider object ambiguity caused by object orientation, occlusion, camera focus, resolution limitations, and lighting conditions. New metrics are needed.

The contributions of this paper are as follows. First, we discuss existing evaluation metrics for relevant computer vision technologies and their limitations when applied to real-life MR Semantic Comprehension. We introduce Dynamic Object Comprehension as a new paradigm that creates a continuity between physical and digital worlds by building visual perception of the physical world over time and space. Finally, we describe a new set of evaluation goals for systems that implement Dynamic Object Comprehension.

## 2. IMAGE CLASSIFICATION METRICS

Metrics allow us to compare the relative quality of solutions, and they essentially define the objectives for solving a target problem. A metric that does not properly embody the desired traits of a solution would motivate the development of suboptimal or wrong solutions. A great example is in the dramatic evolution of image classification solutions that emerged from the ImageNet Competition (ILSVRC) [11-14]. The goal of the competition was to classify images into one of 1000 possible classes. Using the metrics of Top-1 (and Top-5) Error would seem like a reasonable choice. With these metrics, it has been reported that systems have surpassed human-level accuracy in image classification for general purpose [25] and domain-specific [26] applications. Yet, we do not see robust systems in real-life scenarios.

The problem is that Top-1 Error rewards systems that make guesses in the face of uncertainty instead of allowing the system to say that it doesn't know. There is no penalty for being wrong but there is reward for being right. This behaviour is further advantageous for improving the Top-1 Error metric due to the ILSVRC evaluation methodology where test images always belong to one of the 1000 classes. What emerges are systems that carve the input space (the space of all possible input images) into 1000 partitions (a behaviour typically imposed via the SoftMax activation function). When such a system is presented with an image that does not belong to one of the supported classes (this does not occur in the competition but would in real-life scenarios), the system by definition will make a mistake because it can only classify the image into one of the supported classes.

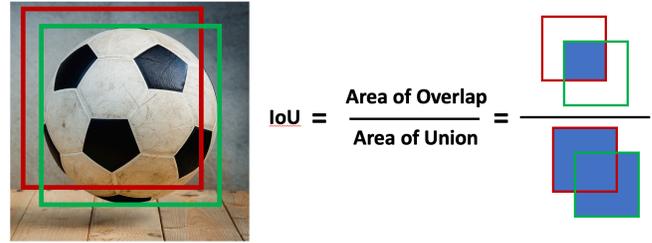

Fig 1. Intersection over Union. Green is the ground-truth bounding-box and red is the predicted bounding-box

Now that these technologies are mature enough to attempt real-life deployment, metrics that reflect real-life requirements are needed. For example, new metrics should encourage systems that can predict that the image does not belong to any of the known classes. We refer to this as an Out-of-Scope (OOS) prediction. To achieve this, we first define the following terms:

- True Positive (TP): When the predicted class matches the ground-truth class, and the class is a known class.
- False Positive (FP): When the predicted class does not match the ground-truth class and the predicted class is a known class.
- False Negative (FN): When the prediction is for OOS, but the ground-truth class is one of the known classes.
- True Negative (TN): When the prediction is for OOS, and the ground-truth class is OOS.

A metric and evaluation framework for the classification aspects of MR Semantic Comprehension should consider TP, FP, FN, TN. Top-1 Error, for example, only considers TP.

## 3. OBJECT DETECTION METRICS

Object Detection seems like a close fit for providing Semantic Comprehension for MR systems. Similar to Image Classification, competitions have been a powerful driver of research in this area. Some notable competitions include PASCAL VOC [27], COCO Detection Challenge [28], Google Open Images [29]. The metrics used by these competitions have directly influenced the course of research.

The goal of an Object Detection system is to find the objects of interest in an image. More specifically, it is common that a detection from such a system be described by a bounding-box (BBox) that encompasses the object of interest within the image, a classification of the object, and a confidence score. Assessing the accuracy of the classification component is similar to the Image Classification problem.

Assessing the accuracy of the BBox prediction requires a definition of correctness. Standard practice is to use the Intersection over Union (IoU). The IoU of a BBox prediction is the ratio of the area of overlap between the predicted BBox and the ground-truth BBox, and the area of the union between the detected BBox and the ground-truth BBox. This is illustrated in Fig 1. If the IoU is above some evaluation

threshold, then the BBox prediction is deemed correct. The following terms can then be defined:
- True Positive (TP): When a BBox with sufficient IoU is predicted for a labelled object in the image
- False Positive (FP): When a BBox is predicted for an object not in the image
- False Negative (FN): When a BBox prediction is not made for a labelled object in the image
- True Negative (TN): When a BBox prediction is not made for an object not in the image.

It is then common to define the Precision and Recall metrics:

$$Precision = \frac{TP}{TP + FP} \qquad Recall = \frac{TP}{TP + FN}$$

The intuition is that by achieving high Recall, the detector would detect all labelled objects (FN=0), and by achieving high Precision the detector would not incorrectly detect non-existent objects (FP=0). To develop a single metric, the widely accepted approach is to use some variant of Average Precision (AP). AP is an evaluation of the Area Under the Curve (AUC) of the Precision/Recall relationship. Further details and examples can be found in [30]. Of notable interest TN is generally not used because it is argued that there are infinite TN scenarios in Object Detection [30]. We believe this to be an important metric, but it is difficult to implement using BBoxes and IoU. Overall, these metrics are not sufficient for evaluating MR Semantic Comprehension.

## 4. OBJECT SEGMENTATION AND OBJECT TRACKING METRICS

Due to space, we will not go into detail about metrics related to Object Segmentation [31] and Object Tracking [18]. The concepts and limitations that will be discussed generally also apply. We refer the reader to the referenced literature for details of those metrics. However, as they share a lot of characteristics with the already discussed metrics, we will discuss their limitations in the following sections as well.

## 5. LIMITATIONS OF CURRENT METRICS AND PROPOSED NEW METRICS

An intuitive approach to MR Semantic Comprehension may be to apply Objected Detection or Object Segmentation to each frame of the MR system's video data. In this context, we discuss how existing evaluation metrics are insufficient.

### 5.1. Acceptable Detection Latency and Accuracy

Using the previously discussed Object Detection accuracy metric means that the ideal solution would achieve perfect detection for all objects on each frame. However, this is generally not required for MR Semantic Comprehension.

We first define a new concept called the Acceptable Detection Latency (ADL). First, Detection Latency is the time between when an object of interest enters the view of the MR system for the first time, until the time that the MR system achieves Semantic Comprehension for the object. ADL is the acceptable Detection Latency. ADL is a function of two human perception components: 1) the latency between when observable change occurs and when the change registers in the mind of the observer [32], and 2) the human expectation for system responsiveness. The latter may be affected by elements such as User Experience and User Interface Design.

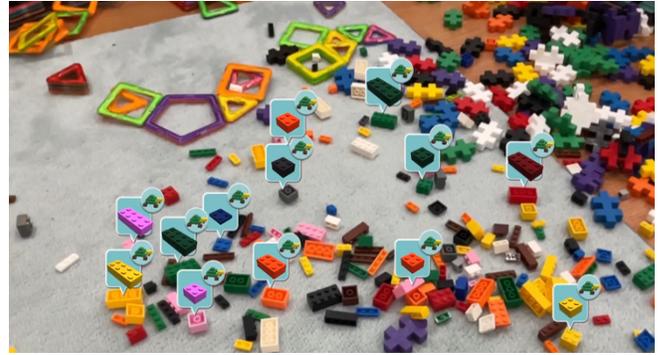

Fig 2. MR System for identifying toy blocks [33]

ADL embodies the requirement for how quickly the system must get to an answer before the user experience is negatively impacted. Consider Fig 2. where an MR System is observing a scene which contains a variety of blocks and tiles from various toy building systems. Consider that the MR System needs to identify the type and 3D location of blocks from one of the building systems as the user looks around the scene. The ADL for this scenario may be on the scale of seconds as opposed to on the scale of the video capture framerate (on the scale of milliseconds).

We can augment an accuracy metric with ADL. Detections no longer need to be complete on each frame. An accurate detection is only required to manifest within the ADL. This unlocks a new trade-off not previously possible. Exploring this new trade-off enabled by ADL is a new area of research.

### 5.2. Ambiguity

Existing metrics fail to consider Ambiguity caused by object orientation, observing perspective, occlusion, camera focus, resolution limitations, and/or lighting conditions. Simply put, under certain conditions, there is not enough information to understand an object. For example, consider the case of Ambiguity caused by observing perspective as illustrated in Fig. 3. We have objects from two supported classes (cube and rectangular cube). When observing the rectangular cube from certain perspectives, it is impossible to determine to which class the object belongs. This is an ambiguous scenario.

In existing frameworks, the ground-truth class label for such a projection may be the rectangular cube class. A system that correctly predicts this label is only doing so by chance and should not be counted as an accurate prediction in the end metric.

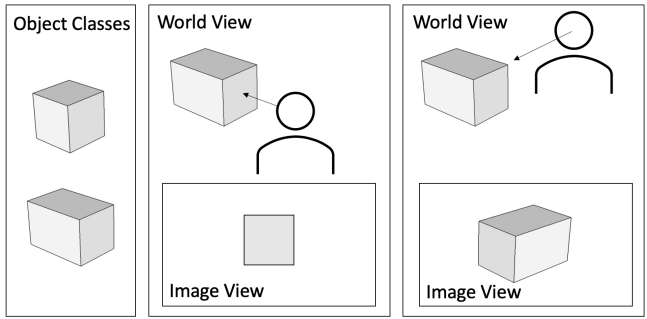

Fig 3. Ambiguity due to observing perspective

### 5.3. Out-of-Scope Predictions

In real-life applications, the number of classes that an MR system must understand is often quite small. The set of classes is often constrained by the goals of the application and/or the environment. Consider again the application from Fig. 2. The number of different block types that need to be understood may be only on the order of 10s or 100s. Furthermore, using such a concrete example, it does not make sense to partition the entire input space into only the class of supported block types. This example further highlights the importance of an OOS prediction. With the concept of OOS, accounting for True-Negatives (a limitation discussed earlier) can be implemented. New metrics that consider the impact of OOS predictions are needed.

### 5.4. Localization and Tracking in 3D

2D BBoxes are the wrong tool. In terms of localization, the main goal of MR Semantic Comprehension is to understand object locations in the 3D world coordinate system, and not where they reside in a 2D projection. The latter is a means and not an end. BBoxes are also assumed to be aligned with the 2D image coordinate axes. But this is arbitrary when considering the MR user's (i.e., the camera's) 6DoF when moving through the world. Consider Fig 4. Which illustrates an MR user with a HMD viewing a scene. The BBox for the laptop changes dramatically with a slight head tilt. It is hard to imagine how the BBoxes in the projected images matter when building MR Semantic Comprehension.

3D BBoxes have been proposed [23] because it is a natural extension to 2D BBoxes. In the case of 3D bboxes, the goal is to understand the 3D position and also the orientation of the object. Many MR applications do not need the orientation information (again one example is shown in Fig. 2). We propose that BBoxes be abandoned completely. If orientation is desired, then a separate method to deliver that information is more appropriate.

Although we did not discuss metrics for Object Tracking research in detail, common metrics for Object Tracking precision (e.g., tracking error) and robustness (e.g., ability to recover from occlusion or when objects move out of view), are of relevance to MR Semantic Comprehension. However, they also suffer from similar limitations of operating in the the 2D image coordinate system and deal with BBoxes [18].

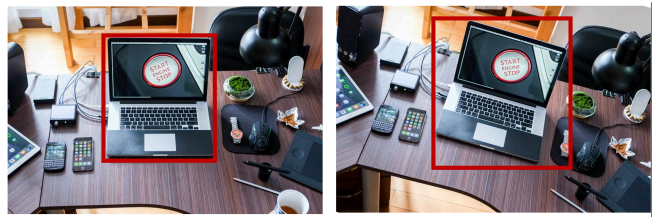

Fig 4. Change in bounding-box due to observing perspective

Since BBoxes are such a fundamental aspect to Object Detection and Tracking, new metrics are needed once we move away from the bounding-box and focus on localization, orientation, and tracking in the 3D world.

## 6. DYNAMIC OBJECT COMPREHENSION

Existing computer vision techniques such as image classification, object detection, object segmentation and object tracking are not directly suitable for providing Semantic Comprehension for MR systems. Aspects of these techniques will no doubt play a role. However, we propose that a new paradigm be defined from the ground-up to address this need without the hindrance of trying to transform other tools that are not quite the perfect fit for the problem.

The goal is to achieve real-time Semantic Comprehension of 3D objects that exist in a 3D world *interpreted* through 2D data. We call this Dynamic Object Comprehension.

More specifically, Dynamic Object Comprehension is the simultaneous, real-time, identification, localization, tracking, and Semantic Comprehension of one or many 3D objects of interest across many object classes, within a 3D environment. The motivating application is Mixed-Reality such as to enable real-time interaction between physical and virtual worlds and unlocking next generation applications.

An example of Dynamic Object Comprehension can be found here [33] and with a specific example of a real operating system here (https://vimeo.com/678457978).

## 7. CONCLUSION

In this paper, we discussed how evaluation metrics for relevant computer vision techniques such as Image Classification, Object Detection, Object Segmentation and Object Tracking are not appropriate for the problem of MR Semantic Comprehension. We described in detail desirable traits of possible new metrics. Furthermore, we propose that we should stop trying to retrofit existing computer vision metrics when trying to evaluate MR Semantic Comprehension problem because this may lead to sub-optimal solutions. Instead, we proposed a new paradigm called Dynamic Object Comprehension with the hope of motivating new research and new solutions, and we invite the community to consider this new line of exploration.